\documentclass[final,5p,times,twocolumn]{elsarticle}

\usepackage{amsmath,amsfonts,amssymb,bm}
\usepackage{array}
\usepackage{graphicx}
\usepackage{xcolor}
\usepackage{soul}
\usepackage{booktabs}
\usepackage{multirow}
\usepackage{url}
\usepackage[breaklinks=true]{hyperref}
\usepackage{algorithm}
\usepackage{algorithmic}

\journal{Information Fusion}

\usepackage{amssymb}
\usepackage{amsmath}

\journal{Information Fusion}
\begin{document}

\begin{frontmatter}



\title{} 

\title{QCC-WM: Linking Macro-State Dynamics and Visual Generation via Quotient Causal Construction}
\author[1]{Zhizhong Fu\fnref{equal}}
\author[1,5,7]{Wei Zhou\fnref{equal}}
\author[2]{Zhaoyang Jiang}
\author[1]{Yudong Lin}
\author[4]{Songlun Li}
\author[5]{Zhi Li}
\author[7]{Feifan Chen}
\author[1]{Gaoming Li}
\author[1]{Xiaorong Ding}
\author[1]{Yifan Chen\corref{cor1}}

\fntext[equal]{Zhizhong Fu and Wei Zhou contributed equally to this work.}

\cortext[cor1]{Corresponding author: Yifan Chen (yifan.chen@uestc.edu.cn).}

\affiliation[1]{organization={School of Life Science and Technology, University of Electronic Science and Technology of China},
            city={Chengdu},
            postcode={610054},
            country={China}}

\affiliation[2]{organization={School of Health and Wellbeing, University of Glasgow},
            city={Glasgow},
            postcode={G12 8TB},
            country={United Kingdom}}

\affiliation[3]{organization={Department of Organ Transplantation, Sichuan Provincial People’s Hospital, University of Electronic Science and Technology of China},
            city={Chengdu},
            country={China}}
\affiliation[4]{organization={School of Information and Software Engineering, University of Electronic Science and Technology of China},
            city={Chengdu},
            postcode={611731},
            country={China}}

\affiliation[5]{organization={Department of Radiology, Huzhou Maternity \& Child Health Care Hospital},
            city={Huzhou},
            postcode={313000},
            country={China}}

\affiliation[6]{organization={Hepatological Surgery Department, Huzhou Central Hospital, Fifth School of Clinical Medicine of Zhejiang Chinese Medical University},
            city={Huzhou},
            postcode={313000},
            country={China}}

\affiliation[7]{organization={Department of Radiology, Huzhou Maternity \& Child Health Care Hospital Affiliated to Huzhou Normal University},
            city={Huzhou},
            postcode={313000},
            country={China}}
\begin{abstract}

A key challenge in action-conditioned world modeling is to construct macro-states from high-dimensional observations that support recursive prediction under actions, while keeping their predicted evolution consistent with visual changes in generated observations. We propose QCC-WM, a visual world model based on Quotient-based Causal Construction. QCC-WM uses action responses to guide macro-variable construction for recursive action-conditioned prediction. A shared reference quotient measurement constrains the consistency between predicted states and observed changes, allowing macro-level predictions to provide structured constraints for future visual generation.
Using translational and rotational actions and their predicted responses, the model aggregates local visual features into macro-nodes with spatial correspondence through causal abstraction, and recursively predicts node content and motion under a shared reference measurement. Macro-level predictions guide future observation generation, while predicted motion transports historical appearance information. Node-level errors between generated observations and macro-level targets are then mapped back to the corresponding spatial regions for correction, improving consistency between visual generation and macro-level predictions.
On the US-VLA dataset, QCC-WM achieves an ROI-MSE of $0.01861\pm0.00008$ for four-step prediction. Compared with the best baseline at each prediction horizon, it reduces LPIPS by $48.59\%$, $41.68\%$, and $30.23\%$ for four-step prediction and eight- and sixteen-step recursive extrapolation, respectively. Ablation studies and mechanism analyses show that motion-guided reference transport mainly improves pixel reconstruction, while node-level error feedback further improves perceptual quality and consistency between generated observations and macro-level predictions.
\end{abstract}


\end{frontmatter}



\section{Introduction}
\label{sec:introduction}

Causal representation learning (CRL) aims to identify high-level variables related to the underlying generative mechanisms from high-dimensional observations, providing representations for analyzing action outcomes, intervention effects, and distribution shifts~\cite{scholkopf2021towards}. For world models, this goal further requires learned states to evolve recursively under actions while maintaining their correspondence with observable changes. However, a fundamental challenge in high-dimensional visual settings is that \textbf{macro-level causal variables suitable for recursive prediction are generally not given and must be constructed from fine-grained observations.} Local features can describe visual appearance and changes, but they do not necessarily form macro-states with consistent measurement definitions and predictable transition rules. Building action-conditioned world models therefore requires addressing both how macro-variables are formed and how they represent state evolution under successive actions.

Research on macro-variable construction and causal abstraction offers theoretical insights into this problem. Chalupka et al. studied how macro-level cause and effect variables can be constructed from the causal behavior of micro-level measurements~\cite{chalupka2016multi}. Hoel et al. showed that, for specific systems and under the effective information measure, suitable macro-level descriptions can exhibit greater causal effectiveness than micro-level descriptions~\cite{hoel2013quantifying}. Rubenstein et al. further emphasized that causal models at different levels should yield consistent predictions under explicitly defined intervention mappings~\cite{rubenstein2017causal}. These studies suggest that micro-to-macro mappings should be designed around the mechanisms and dynamics they preserve. For world models, we therefore consider a dynamics-oriented abstraction principle: \textbf{macro-variable construction should be aligned with future prediction so that abstract state changes correspond to evolution at the observation level.}

Existing visual world models have explored various state representations and prediction methods. DINO-WM learns action-conditioned dynamics in pretrained spatial features~\cite{zhou2025dinowm}, while IRASim incorporates action trajectories into a diffusion model to generate future observations~\cite{zhu2025irasim}. For structured modeling, SlotFormer learns recursive dynamics over object-level representations~\cite{wu2023slotformer}, and Slot-JEPA organizes prediction through semantic slots and spatial masks~\cite{wen2026slotjepa}. Motion-I2V uses predicted motion to transport reference features, connecting motion modeling with visual generation~\cite{shi2024motioni2v}. These studies provide important foundations for state prediction, structured representations, and spatially organized generation. Recursive macro-state prediction nevertheless raises several questions: whether nodes have comparable measurement definitions across time steps, whether predicted state changes are constrained by observed evolution, and whether macro-level predictions retain the correspondence needed to map them back to observation space. These questions directly affect the preservation of state meaning and the accumulation of prediction errors during recursive prediction.

To address these issues, we introduce
\textbf{QCC (Quotient-based Causal Construction)}
and develop the action-conditioned world model QCC-WM.
QCC uses action-guided soft quotient mappings to aggregate local visual features into macro-nodes with spatial support.
It recursively predicts content and motion states through inter-node dependencies and action-conditioned transitions.
Within each prediction window, the model fixes the reference quotient mapping and its alignment,
and constrains macro-state prediction through supervision on states and their temporal increments.
This encourages consistency with observed evolution under the same quotient measurement.

To translate macro-level predictions into observation generation, the model uses predicted motion to transport reference features and latent variables,
construct observable macro-level targets, and correct the initial generation.
It then reads back the generated observations through the same quotient measurement,
computes node-level errors relative to the macro-level targets,
and maps these errors back to local latent space through the node--space correspondence.
The reference quotient mapping thus connects node construction, macro-state prediction, and spatial correction,
linking macro-level dynamics with future observation generation.

Our main contributions are as follows:
\begin{enumerate}
    \item \textbf{An action-guided quotient-based causal construction mechanism, QCC.}
    We construct macro-nodes with spatial support through soft quotient mappings,
    discover a macro-level causal structure of temporal dependencies among nodes under a shared reference measurement,
    and use this structure to organize action-conditioned dynamics.
    This connects macro-node construction and structure discovery with recursive prediction of content and motion states.

    \item \textbf{A macro-level dynamics learning method under a fixed reference quotient measurement.}
    We fix the reference quotient mapping and content alignment within each prediction window
    to provide consistent measurement definitions for macro-states across time steps.
    Supervision on states and their temporal increments constrains recursive content and motion prediction,
    encouraging consistency between macro-level predictions and observed evolution under the same reference measurement.

    \item \textbf{A generation correction interface combining motion transport with readback under the same quotient measurement.}
    We use predicted motion to transport historical appearance information,
    then map node-level errors obtained through readback to the generation latent space through spatial correspondence,
    improving consistency between future observations and macro-level predictions.
\end{enumerate}

The remainder of this paper is organized as follows.
Section \ref{sec:related_work} reviews related work on causal representation learning, structured world models, and motion-guided generation.
Section \ref{sec:method} presents QCC-WM, including quotient-based causal construction, macro-state prediction, motion-guided reference transport, spatial correction, and staged optimization.
Section \ref{sec:results} describes the experimental setup and evaluates model performance and module contributions through comparative experiments, ablation studies, and mechanism analyses.
Section \ref{sec:conclusion} concludes the paper and discusses future directions.

\section{Related Work}
\label{sec:related_work}

\subsection{Causal Representation Learning and Macro-Variable Construction}

Causal representation learning seeks to identify latent variables related to generative mechanisms from high-dimensional observations, enabling mechanism analysis, intervention prediction, and generalization across distributions~\cite{scholkopf2021towards}. In dynamic visual settings, CITRIS uses temporal observations and known intervention targets to identify latent causal factors under specific assumptions~\cite{lippe2022citris}. BISCUIT studies causal variable learning with unknown binary interaction mechanisms and establishes corresponding identifiability conditions~\cite{lippe2023biscuit}. These studies show that causal interpretations depend closely on additional supervision and structural assumptions. QCC focuses on constructing macro-nodes for action-conditioned prediction. Its node measurements and dependency structures are learned and evaluated as predictive interfaces within the model; action inputs and interventions within the model alone do not provide such identifiability guarantees.

Macro-variable construction also concerns relationships between different levels of description. Chalupka et al. used causal partitions to connect micro-level measurements with macro-level cause and effect variables~\cite{chalupka2016multi}. Hoel et al. analyzed macro-level causal effectiveness in specific systems using effective information~\cite{hoel2013quantifying}. Rubenstein et al. formalized intervention consistency across levels through exact transformations between structural equation models~\cite{rubenstein2017causal}. In reinforcement learning, DeepMDP learns state representations by predicting rewards and next-state distributions in latent space, and establishes connections to bisimulation~\cite{gelada2019deepmdp}. These studies provide a basis for abstractions that preserve underlying dynamics. We apply this idea to spatial aggregation within visual observations: action-guided soft quotient mappings define macro-level measurements, whose future evolution is constrained to match observed changes under the same reference mapping. This spatial quotient construction addresses a different modeling object from full environment-state abstraction, and the prediction consistency imposed here does not imply strict intervention consistency.

\subsection{Action-Conditioned Dynamics and Structured World Models}

Latent world models support future prediction by learning transition rules in compact state spaces. PlaNet combines deterministic and stochastic transitions and introduces latent overshooting as a multi-step variational objective~\cite{hafner2019planet}. Dreamer optimizes behavior over imagined latent trajectories, using recursive state prediction for control~\cite{hafner2020dreamer}. DINO-WM learns action-conditioned patch dynamics from pretrained spatial features and performs planning through goal-feature prediction~\cite{zhou2025dinowm}. These methods connect latent state prediction with downstream decision-making. QCC further addresses how local observations are organized into macro-level prediction units. A shared reference measurement constrains node states and their temporal increments, providing comparable observation-based targets for recursive prediction in the abstract state space.

Structured world models explicitly organize prediction units within a scene. DINOSAUR learns object-level representations by reconstructing self-supervised pretrained features~\cite{seitzer2023dinosaur}. SlotFormer learns autoregressive dynamics over slots to predict future object states~\cite{wu2023slotformer}. Slot-JEPA decomposes dense features into semantic slots and time-varying spatial masks, combining transition memory and transport objectives for prediction and planning~\cite{wen2026slotjepa}. These methods establish foundations for feature grouping, structured recursive prediction, and spatial correspondence. Our specific design incorporates action plans into quotient mapping construction and uses the same reference quotient mapping for macro-state measurement, future-state supervision, and mapping predictions back to spatial locations. QCC nodes are defined as action-conditioned aggregate measurements, whose predictive value must be assessed through appropriate controlled comparisons.

\subsection{Generative World Models and Motion-Guided Spatial Feedback}

Generative world models synthesize future observations from actions or structured states. IRASim models robotic interaction videos using a diffusion Transformer with frame-wise action conditioning~\cite{zhu2025irasim}. To connect structured representations with generation, SlotDiffusion uses a slot-conditioned latent diffusion decoder and applies the learned representations to object dynamics modeling~\cite{wu2023slotdiffusion}. SlotDiT generates future slot trajectories through text-conditioned diffusion and then decodes them into images~\cite{plepi2026slotdit}. QCC uses macro-level content and motion as states for recursive prediction and converts these predictions into generation conditions in image latent space through explicit quotient mappings. Its focus is the interface between macro-level dynamics and observation generation.

Explicit motion modeling provides another way to connect state changes with visual synthesis. Motion-I2V first predicts pixel trajectories and then propagates reference image features through motion-enhanced temporal attention~\cite{shi2024motioni2v}. FLIP combines flow generation, flow-conditioned video generation, and value modeling for generative planning in visual space~\cite{gao2024flip}. DeXplicit predicts explicit 3D states and deformations from actions and transports appearance information to guide video generation~\cite{lee2026dexplicit}. These studies provide direct precedents for motion prediction and reference content transport. We use a learned visual quotient space as an intermediate representation, where regional motion and content changes arise from recursive macro-state prediction and are mapped back to local space through the retained quotient mapping. Spatial feedback thus forms a prediction pipeline together with QCC node measurements and recursive dynamics. Its contribution is assessed through generation evaluation and interventions on the model interfaces.

\section{Methods}
\label{sec:method}

\subsection{Problem Formulation and QCC Overview}
\label{sec:method_setup}

Given historical visual observations $X=I_{-L+1:0}$ and a known action plan $A=(a_0,\ldots,a_{H-1})$, we learn the conditional distribution $p(Y\mid X,A)$ of future observations $Y=I_{1:H}$.
The central problem is to construct macro-states that support action-conditioned prediction while preserving their connection to observation space, allowing their predictions to guide future generation.

We propose \textbf{QCC (Quotient-based Causal Construction)}, which uses action-guided soft quotient mappings to organize local visual representations into macro-nodes with explicit spatial origins.
QCC jointly defines the aggregation sources, reference measurements, and spatial correspondence of nodes, and establishes action-conditioned transitions and inter-node dependencies over the constructed macro-variables.
We refer to these state units as macro-level causal nodes, which represent action-driven evolution and inter-node interactions within the model.

\begin{figure*}[t]
    \centering
    \includegraphics[width=\textwidth]{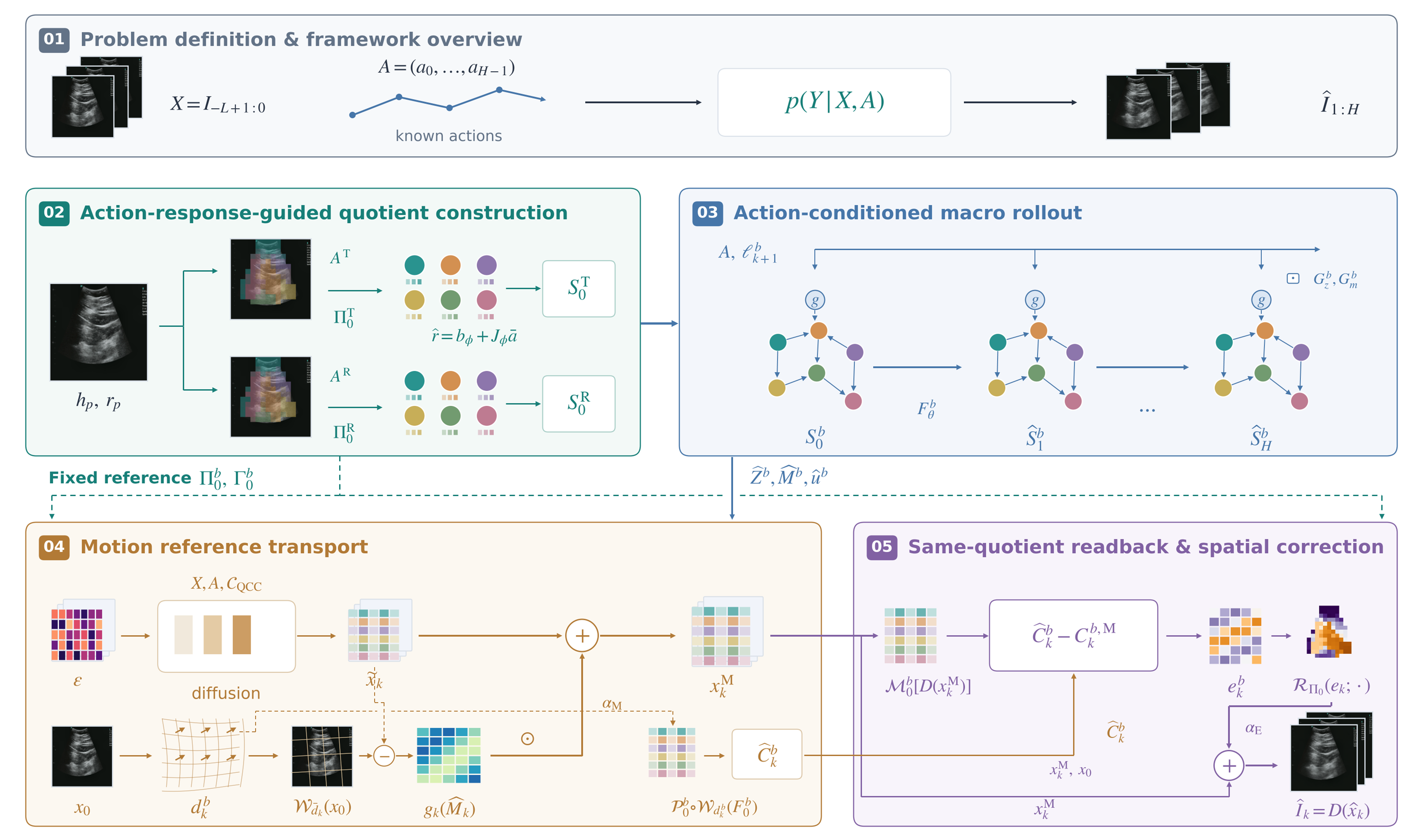}
    \caption{Overview of QCC-WM. QCC first constructs macro-nodes with spatial origins and reference measurements. Action-conditioned dynamics recursively predict content and motion states. Predicted motion transports reference features and latent variables to construct observable macro-level targets and perform motion-guided reference correction, respectively. The model then reads back the current generation under the same quotient measurement and maps the remaining node-level errors to the corresponding spatial locations to obtain future observations. The reference quotient mapping connects node construction, macro-state prediction, and spatial generation.}
    \label{fig:qcc_method_overview}
\end{figure*}

As shown in Fig.~\ref{fig:qcc_method_overview}, the framework consists of four components: quotient-based causal construction, action-conditioned macro-state prediction, motion-guided reference transport, and readback under the same quotient measurement followed by spatial correction:
\begin{equation}
\label{eq:qcc_overview}
\begin{aligned}
    &(\Pi_0^b,\Gamma_0^b,S_0^b)
    = Q_\phi^b(X,A^b), \\[3pt]
    &\widehat S_{k+1}^b
    = F_\theta^b\!\left(
        \widehat S_k^b,a_k,A^b,\ell_{k+1}^b;
        G_z^b,G_m^b
    \right), \\[3pt]
    &\widetilde x_{1:H}
    \sim p_\psi\!\left(
        x_{1:H}\mid X,A,\mathcal C_{\mathrm{QCC}}
    \right), \\[3pt]
    &\widehat I_k
    = D\!\left(
        x_k^{\mathrm M}
        +\alpha_{\mathrm E}\mathcal R_{\Pi_0}(e_k;\cdot)
    \right),
\end{aligned}
\end{equation}
where $b\in\mathcal B=\{\mathrm T,\mathrm R\}$ denotes the quotient construction branch guided by translational or rotational action components, and $A^b$ is the corresponding action guidance.
Both dynamics models and the generator retain the complete action input.
$\Pi_0^b$ and $\Gamma_0^b$ denote the reference soft assignment and reference content alignment, respectively, while $G_z^b,G_m^b$ specify the dependency supports for the content and motion channels.
$S_k^b=(Z_k^b,M_k^b,g_k^{z,b},g_k^{m,b})$ contains node content, node motion memory, and their corresponding shared states; $\ell_{k+1}^b$ is the node control derived from the action responses predicted by QCC.
$\mathcal C_{\mathrm{QCC}}$ denotes the generation conditions produced by macro-state prediction, $\widetilde x_k$ is the initially generated latent variable, $x_k^{\mathrm M}$ is the latent variable after motion-guided reference correction, $e_k$ is the remaining node-level error obtained through readback under the same quotient measurement, and $D$ is the observation decoder.

\subsection{Action-Response-Guided Quotient-Based Causal Construction}
\label{sec:method_quotient}

We use the pretrained visual encoder
$E_{\mathrm{obs}}$ of I-JEPA~\cite{Assran_2023_CVPR}
to extract observation representations and obtain frame-wise content features through branch-specific mappings:
$F_k^b=\Phi^b(E_{\mathrm{obs}}(I_k))$.
Historical features and their local temporal differences are used to organize nodes and initialize motion,
while frame-wise content features measure node states at each time step.
Each branch maintains $K$ global prototypes
$\{c_j^b\}_{j=1}^{K}$, with prototype indices identifying the nodes.
QCC uses the action plan to modulate the affinity between historical local features and prototypes.
Combined with spatial distances, these affinities determine soft assignments
$\Pi_{pj}^{b,\mathrm{proto}}$ within the valid observation region.
The assignments are normalized across nodes and refined by iteratively updating node centroids,
thereby defining the local spatial support of each node.

\paragraph{Action-response-driven quotient refinement.}
To connect node organization with potential action-induced changes, QCC further predicts local action responses from historical descriptors.
For future time step $k$, let $\bar a_k^b=\sum_{r=0}^{k-1}a_r^b$ denote the cumulative action guidance:
\begin{equation}
    \widehat r_{k,p}^b
        =b_{\phi,k}^b(h_p^b)
          +J_{\phi,k}^b(h_p^b,r_p)\bar a_k^b,
    \label{eq:qcc_local_action_response}
\end{equation}
where $b_{\phi,k}^b$ is the response base term conditioned on history, and $J_{\phi,k}^b$ parameterizes the interaction between local mechanisms and actions.
The response includes cumulative motion and appearance changes relative to the reference latent variable, connecting feature grouping with the future changes to be represented.

We subtract the spatial mean of the action response term over the valid region, normalize it using training statistics, and concatenate it across time to form the local response descriptor $\chi_p^b$.
This descriptor compares relative responses across locations under the current action plan.
Starting from the prototype grouping, QCC sequentially refines spatial assignments using history-conditioned response encodings and explicit mechanism--action response encodings.
Both refinements alternate between response aggregation and assignment updates.
For each refinement, let $\Pi^{b,\mathrm{prior}}$ denote the input assignment and $\chi^b$ the corresponding response descriptor:
\begin{align}
    &\bar\chi_j^{b,(n)}
        =\frac{\sum_p\Pi_{pj}^{b,(n)}\chi_p^b}
                {\sum_p\Pi_{pj}^{b,(n)}+\epsilon}, \nonumber\\
    &\widetilde\Pi_{pj}^{b,(n+1)}
        \propto m_p\exp\!\left[
            \log\!\left(\Pi_{pj}^{b,\mathrm{prior}}+\epsilon\right)
            -\lambda_r\|\chi_p^b-\bar\chi_j^{b,(n)}\|_2^2
          \right], \nonumber\\
   & \Pi_{pj}^{b,\mathrm{new}}
        =(1-\zeta_b)\Pi_{pj}^{b,\mathrm{prior}}
          +\zeta_b\widetilde\Pi_{pj}^b,
    \label{eq:qcc_response_quotient}
\end{align}
where $\zeta_b$ is a bounded strength learned separately for each refinement layer, and the final assignment is denoted by $\Pi_0^b$.
Node assignments are thus jointly constrained by historical spatial structure and predicted action responses, organizing local evolution information while preserving prototype identities.
At inference, action responses are predicted entirely from history and the known action plan; observed future responses are used only for training supervision.

\paragraph{Reference quotient measurement and node initialization.}
QCC further uses historical descriptors and predicted action responses to produce a bounded reference content alignment $\Gamma_0^b$, which remains fixed within the current prediction window.
We define the normalized aggregation weights as
\begin{equation}
    \omega_{0,pj}^b
        =\frac{\Pi_{0,pj}^b}
                {\sum_q\Pi_{0,qj}^b+\epsilon},
    \qquad 
    \\
    \mathcal P_{\Pi_0^b,\Gamma_0^b}(F)_j
        =\sum_p\omega_{0,pj}^b\,
            \bigl(\Gamma_{0,p}^b\odot F_p\bigr).
    \label{eq:macro_measurement}
\end{equation}
The corresponding reference quotient measurement is
\begin{equation}
    \mathcal M_0^b(I)
        =\mathcal P_{\Pi_0^b,\Gamma_0^b}
          \!\left(\Phi^b(E_{\mathrm{obs}}(I))\right),
    \qquad
    Z_{k}^{b,\star}=\mathcal M_0^b(I_k).
    \label{eq:qcc_reference_measurement}
\end{equation}
This definition unifies spatial assignment, content alignment, and frame-wise measurement.
The initial node content is set to $Z_0^b=Z_0^{b,\star}$.
Motion memory is initialized by quotient aggregation of historical content changes, estimated historical optical flow, and local mechanism descriptors.
Reference centroids are computed from the final assignment $\Pi_0^b$.
Shared content and motion states are obtained through global aggregation over the valid region, providing a reference for overall changes to the local nodes.

\subsection{Action-Conditioned Macro-State Prediction under a Fixed Quotient Measurement}
\label{sec:method_dynamics}

\paragraph{Node dependencies and step-wise response control.}
We construct candidate node dependency supports through temporal dependency analysis of macro-level measurements from training observations, incorporating action and reference conditions~\cite{runge2019detecting}.
The analysis controls for the complete action plan, reference support, and past measurements of shared states.
Fixed directed supports $G_z^b,G_m^b$ organize inter-node messages for the content and motion channels.
For channel $c\in\{z,m\}$, incoming edge weights are defined as
\begin{equation}
    B_{ij}^{c,b}
        =\frac{G_{c,ij}^b\operatorname{clip}(\gamma_{ij}^{c,b},0,1)}
               {1+\sum_lG_{c,lj}^b
                    \operatorname{clip}(\gamma_{lj}^{c,b},0,1)},
    \label{eq:message_normalization}
\end{equation}
where $\gamma_{ij}^{c,b}$ is a learnable strength within the support, and the direction from $i$ to $j$ represents information transfer from node $i$ at the previous time step to node $j$ at the current time step.
The directed supports remain fixed during recursive prediction, with content and motion using their respective message weights.

The action responses predicted by QCC also enter the recursive dynamics through step-wise node controls.
We take temporal differences of the cumulative motion responses in Eq.~\eqref{eq:qcc_local_action_response} and aggregate them under the reference quotient mapping:
\begin{equation}
    \ell_{k,j}^b
        =\sum_p\omega_{0,pj}^b\,
            \psi_\ell^b\!\left(
                \widehat r_{k,p,\mathrm{mot}}^b
                -\widehat r_{k-1,p,\mathrm{mot}}^b
            \right),
    \quad \widehat r_{0,p,\mathrm{mot}}^b=0,
    \label{eq:qcc_persistent_control}
\end{equation}
where $\psi_\ell^b$ is a bounded response encoding function.
This control allows local action responses learned during construction to contribute to node updates at every step, jointly driving macro-state evolution with the complete action input.

\paragraph{Recursive evolution of shared and local states.}
To separate overall changes from local differences, we first update the shared states:
\begin{equation}
    (g_{k+1}^{z,b},g_{k+1}^{m,b})
        =F_g^b(g_k^{z,b},g_k^{m,b},a_k,A^b),
    \label{eq:shared_dynamics}
\end{equation}
and define local residuals as $\widetilde Z_{k,j}^b=Z_{k,j}^b-g_k^{z,b}$ and
$\widetilde M_{k,j}^b=M_{k,j}^b-g_k^{m,b}$.
Content messages combine node content, motion, reference positions, and fixed historical context, while motion messages are extracted from motion residuals:
\begin{align}
    \mathsf m_{k,j}^{z,b}
        &=\sum_iB_{ij}^{z,b}\,
            V_z^b\!\left(\widetilde Z_{k,i}^b,
                         \widetilde M_{k,i}^b,\mu_i^b;X\right), \nonumber\\
    \mathsf m_{k,j}^{m,b}
        &=\sum_iB_{ij}^{m,b}\,
            V_m^b(\widetilde M_{k,i}^b),
    \label{eq:node_messages}
\end{align}
where $V_z^b,V_m^b$ denote the message projections for the respective channels.
Inter-node messages use the states at the start of each step, while each node continues to evolve within its residual update blocks.

Node content is updated through a residual mapping, while motion memory is updated through a gated recurrent unit:
\begin{align}
    \widetilde Z_{k+1,j}^b
        &=\mathcal F_z^b\!\left(
            \widetilde Z_{k,j}^b,\widetilde M_{k,j}^b,
            \mathsf m_{k,j}^{z,b},a_k,A^b,
            \ell_{k+1,j}^b,g_k^b,\mu_j^b;X
          \right), \nonumber\\
    Z_{k+1,j}^b
        &=\widetilde Z_{k+1,j}^b+g_{k+1}^{z,b}, \nonumber\\
    \widetilde M_{k+1,j}^b
        &=\operatorname{GRU}^b\!\left(
            \widetilde M_{k,j}^b,
            \Psi_m^b\!\left(
                \widetilde Z_{k+1,j}^b,\mathsf m_{k,j}^{m,b},
                a_k,\ell_{k+1,j}^b,\mu_j^b
            \right)
          \right), \nonumber\\
    M_{k+1,j}^b
        &=\widetilde M_{k+1,j}^b+g_{k+1}^{m,b}, \nonumber\\
    \widehat u_{k+1,j}^b
        &=\rho_m^b(\widetilde M_{k+1,j}^b)
          +\rho_g^b(g_{k+1}^{m,b}),
    \label{eq:node_recurrence}
\end{align}
where $g_k^b=[g_k^{z,b};g_k^{m,b}]$, $\widehat u_{k+1,j}^b$ is the node motion response, and $\mathcal F_z^b$ is the parameterized residual update mapping for node content in branch $b$.
This update combines each node's history, dependency messages, shared changes, and local action responses to recursively predict future sequences from the same initial macro-state without future observations.

Content dynamics are learned under a fixed reference quotient measurement, connecting recursively predicted states with corresponding observed changes:
\begin{equation}
    \operatorname{pr}_zF_\theta^{b,(k)}(S_0^b,A)
        \approx\mathcal M_0^b(I_k),
    \label{eq:quotient_commuting_target}
\end{equation}
where $\operatorname{pr}_z$ extracts node content.
Initial and future observations are measured using the same $\Pi_0^b,\Gamma_0^b$, providing a common reference across prediction steps.
Supervision is applied to both states and their temporal increments:
\begin{align}
    \mathcal L_{\mathrm Q}
        =\Big\langle\,
            &\lambda_z\operatorname{MSE}(\widehat Z^b,Z^{b,\star})
            +\lambda_n\operatorname{MSE}\!\left(
                \operatorname{LN}(\widehat Z^b),
                \operatorname{LN}(Z^{b,\star})
              \right) \nonumber\\
            &+\lambda_\Delta\operatorname{MSE}\!\left(
                \Delta\widehat Z^b,\Delta Z^{b,\star}
              \right)
          \,\Big\rangle_b,
    \quad\Delta Z_k=Z_k-Z_{k-1},
    \label{eq:quotient_consistency}
\end{align}
where $\langle\cdot\rangle_b$ denotes averaging over branches, and MSE is averaged over samples, prediction steps, nodes, and channels.
This constraint aligns macro-states and their changes under a fixed observation reference, providing state sequences with a common measurement basis for subsequent spatial generation.

\subsection{Quotient-Mapping-Based Motion-Guided Reference Transport}
\label{sec:method_transport}

QCC first uses spatial correspondence to condition the initial generation on macro-states.
The model performs frame-level and temporal readouts of node content and motion, and injects node content increments and motion geometry into the generator's local features through the reference quotient mapping.
For example, the spatial field corresponding to node content changes is
\begin{equation}
    \Delta F_k^b(p)
        =\sum_j\Pi_{0,pj}^b
            (\widehat Z_{k,j}^b-Z_{0,j}^b).
    \label{eq:quotient_content_return}
\end{equation}
Together with macro-level motion conditions, this field forms $\mathcal C_{\mathrm{QCC}}$ and guides the generation of the initial latent block $\widetilde x_{1:H}$, while retaining historical visual information and direct action conditioning.

To further translate macro-level motion into changes in observation space, we use the same reference assignment to recover local displacement fields and accumulate them over prediction steps:
\begin{equation}
    \widehat v_k^b(p)
        =\sigma_f\sum_j\Pi_{0,pj}^b\widehat u_{k,j}^b,
    \qquad
    d_k^b(p)=\sum_{r=1}^{k}\widehat v_r^b(p),
    \label{eq:qcc_dense_motion}
\end{equation}
where $\sigma_f$ converts normalized motion responses to the displacement scale required for transport.
We define the reference transport operator through bilinear backward sampling as
$\mathcal W_{d}(F)(p)=F(p-d(p))$,
with displacements interpolated and rescaled to the corresponding feature grid.

Reference information is transported separately in content feature space and generation latent space.
For branch $b$, the reference content features $F_0^b$ are transported using the predicted displacement and measured through the fixed quotient interface to form an observable macro-level target:
\begin{equation}
    \widehat C_k^b
        =\mathcal P_{\Pi_0^b,\Gamma_0^b}
           \!\left(\mathcal W_{d_k^b}(F_0^b)\right),
    \label{eq:qcc_observable_target}
\end{equation}
where $\widehat C_k^b$ represents the observable changes induced by predicted motion in the reference content and is used for comparison with generated observations under the same quotient measurement.
The recursively predicted latent state $\widehat Z_k^b$ supports macro-state prediction and generation conditioning.
The two quantities thus serve latent prediction and observable correction, respectively.

In latent space, we average the displacement fields from the two branches to obtain $\bar d_k$ and transport the latent variable $x_0$ of the last historical observation:
\begin{equation}
    \bar d_k=\frac{1}{|\mathcal B|}\sum_b d_k^b,
    \qquad x_k^{\mathrm{ref}}=\mathcal W_{\bar d_k}(x_0).
    \label{eq:qcc_latent_transport}
\end{equation}
QCC also converts node motion memory into a spatial gate:
\begin{equation}
    g_k(p)
        =\kappa\,\operatorname{Interp}\!\left[
            \frac{1}{|\mathcal B|}\sum_b\sum_j
                \Pi_{0,pj}^b\,sigma\!\left(h_m^b(\widehat M_{k,j}^b)\right)
          \right],
    \label{eq:qcc_motion_gate}
\end{equation}
where $h_m^b$ is the node motion gating network, $\sigma$ is the sigmoid function, and $\kappa$ is a bounded strength.
The gate depends on predicted motion memory and is expanded over the reference spatial support of each node.
Motion-guided reference correction is defined as
\begin{equation}
    x_k^{\mathrm M}
        =\widetilde x_k+
          \alpha_{\mathrm M}g_k\odot
          (x_k^{\mathrm{ref}}-\widetilde x_k),
    \label{eq:qcc_motion_reference_correction}
\end{equation}
where $\widetilde x_k$ is the initially generated latent variable,
$x_k^{\mathrm{ref}}$ is the reference latent variable transported using predicted motion,
$g_k$ contains spatial gating weights, $\alpha_{\mathrm M}$ controls the overall correction strength,
and $\odot$ denotes element-wise multiplication.
The difference between the reference latent variable and the initial generation defines the correction direction.
The spatial gate controls the correction magnitude at each location, incorporating transported historical appearance information into the generation.

\subsection{Readback under the Same Quotient Measurement and Spatial Correction of Node Errors}
\label{sec:method_return}

After motion-guided reference transport, QCC uses the same reference quotient measurement to evaluate the discrepancy between the current generation and the observable macro-level target.
It first decodes the motion-corrected latent variable and reads back its macro-level content:
\begin{equation}
    C_k^{b,\mathrm M}
        =\mathcal M_0^b\!\left(D(x_k^{\mathrm M})\right),
    \qquad e_k^b=\widehat C_k^b-C_k^{b,\mathrm M}.
    \label{eq:qcc_generated_readback}
\end{equation}
Because target construction and generation readback share $\Pi_0^b,\Gamma_0^b$, $e_k^b$ has fixed node definitions and spatial origins.
It represents changes that the current generation has not yet captured under the macro-level measurement.
Readback follows motion-guided reference correction, allowing error correction to operate on generation results that have already been adjusted using reference motion.

Spatial feedback expands errors through the node--location relationships defined by QCC.
One path directly projects node errors, while the other modulates error representations using predicted content and content read back from the generation:
\begin{align}
    &E_k^{b,\mathrm{base}}(p)
        =\sum_j\Pi_{0,pj}^b W_e^b e_{k,j}^b, \nonumber\\
    &r_{k,j}^b
        =W_r^b e_{k,j}^b\odot
            \left[1+\xi^b\!\left(
                [\widehat C_{k,j}^b;C_{k,j}^{b,\mathrm M}]
            \right)\right], \nonumber\\
    &E_k^{b,\mathrm{ctx}}(p)
        =\sum_j\Pi_{0,pj}^b r_{k,j}^b,
    \label{eq:qcc_error_fields}
\end{align}
where $W_e^b,W_r^b$ are channel projections and $\xi^b$ is a node content modulation function.
The error fields are interpolated to the generation latent grid and combined with local context from the current and reference latent variables to produce spatial corrections:
\begin{align}
    \delta x_k^{\mathrm E}
        &=\mathcal D_{\mathrm{base}}\!\left(
            \{E_k^{b,\mathrm{base}}\}_b;x_k^{\mathrm M}
          \right)
          +\mathcal D_{\mathrm{ctx}}\!\left(
            \{E_k^{b,\mathrm{ctx}}\}_b;x_k^{\mathrm M},x_0
          \right), \nonumber\\
    \widehat x_k
        &=x_k^{\mathrm M}+\alpha_{\mathrm E}\delta x_k^{\mathrm E},
    \qquad \widehat I_k=D(\widehat x_k),
    \label{eq:qcc_spatial_error_return}
\end{align}
where $\mathcal D_{\mathrm{base}}$ and $\mathcal D_{\mathrm{ctx}}$ include branch fusion, spatial interpolation, and local context modulation, and are collectively denoted by $\mathcal R_{\Pi_0}$.
This interface converts macro-level errors into generation adjustments with explicit spatial correspondence.

\subsection{Learning Objectives and Staged Optimization}
\label{sec:method_training}

The model is trained for quotient construction, macro-level content prediction, conditional generation, and spatial correction.
The main objective is summarized as
\begin{equation}
\begin{aligned}
\mathcal L_{\mathrm{main}}
={}&\lambda_{\mathrm{group}}\mathcal L_{\mathrm{group}}
  +\lambda_{\mathrm Q}\mathcal L_{\mathrm Q}\\
 &+\lambda_{\mathrm{gen}}\mathcal L_{\mathrm{gen}}
  +\lambda_{\mathrm{ret}}\mathcal L_{\mathrm{ret}},
\end{aligned}
\label{eq:qcc_main_objective}
\end{equation}
where $\mathcal L_{\mathrm{group}}$ encourages quotient construction to preserve observed response information,
$\mathcal L_{\mathrm Q}$ supervises macro-level content prediction under a fixed reference measurement,
$\mathcal L_{\mathrm{gen}}$ trains conditional diffusion generation,
and $\mathcal L_{\mathrm{ret}}$ trains spatial correction.
Training also includes auxiliary supervision on local responses, node and local motion, shared states, and transport alignment, as detailed below.
These objectives are applied to the corresponding modules in separate training stages.
The spatial correction module is trained separately after the core model is frozen.

Local response prediction is supervised using cumulative optical flow and changes relative to the reference latent variable from training observations.
Let $r_k^\star$ denote the observed response normalized using training statistics.
We supervise Eq.~\eqref{eq:qcc_local_action_response} with both response prediction error and local response error after removing the spatial mean.
Response reconstruction after node aggregation further encourages quotient construction to preserve information relevant to future evolution:
\begin{equation}
    \bar r_{k,j}^{b,\star}
        =\sum_p\omega_{0,pj}^b r_{k,p}^\star,
    \\
    \mathcal L_{\mathrm{group}}
        =\Big\langle\,
            \big\|\Pi_0^b\bar r^{b,\star}-r^\star\big\|_\Omega^2
          \,\Big\rangle_b
          +\lambda_{\mathrm{reg}}\mathcal L_{\mathrm{reg}},
    \label{eq:calibration_objective}
\end{equation}
where $\Omega$ denotes the valid support, and $\mathcal L_{\mathrm{reg}}$ regularizes node occupancy and near-duplicate normalized assignment columns.
These objectives connect node organization with observed responses during training, while construction itself uses only history and actions.

Content learning in the macro-level dynamics uses the quotient consistency objective in Eq.~\eqref{eq:quotient_consistency}.
Node motion targets are obtained by aggregating observed optical flow $f_k$ under the same reference assignment:
\begin{equation}
    u_{k,j}^{b,\star}
        =\sum_p\omega_{0,pj}^b f_{k,p},
    \qquad
    \widehat f_{k,p}^b
        =\sum_j\Pi_{0,pj}^b\widehat u_{k,j}^b,
    \label{eq:qcc_motion_supervision}
\end{equation}
where predictions and targets use the same motion scale.
The corresponding losses compare node motion $\widehat u^b$ with $u^{b,\star}$ and reconstructed local fields $\widehat f^b$ with $f$, while also supervising shared content increments and shared motion.
Construction and response adaptation additionally use local motion errors and alignment errors between transported references and true future latent variables to coordinate response control, macro-level motion, and reference transport.

Initial observations are generated in a pretrained visual latent space.
Let $x_k=E_{\mathrm{vae}}(I_k)$.
We construct diffusion targets using the reference residual $r_k^x=(x_k-x_0)/\sigma_x$:
\begin{equation}
    x_{s,k}=\sqrt{\bar\alpha_s}\,r_k^x
             +\sqrt{1-\bar\alpha_s}\,\epsilon_k,
    \qquad
    v_{s,k}^{\star}=\sqrt{\bar\alpha_s}\,\epsilon_k
             -\sqrt{1-\bar\alpha_s}\,r_k^x.
    \label{eq:qcc_diffusion_target}
\end{equation}
Conditioned on history, complete actions, and $\mathcal C_{\mathrm{QCC}}$, the generator minimizes a denoising objective weighted by the valid region:
\begin{equation}
    \mathcal L_{\mathrm{gen}}
        =\frac{\mathbb E\!\left[
            w(p)\bigl(\widehat v_{\psi,s,k}(p,c)-v_{s,k}^{\star}(p,c)\bigr)^2
          \right]}{\mathbb E[w(p)]},
    \label{eq:generation_objective}
\end{equation}
where $w(p)$ is the valid observation region weight, and the expectation averages over samples, diffusion steps, future steps, positions, and channels $c$.

The spatial correction module is trained using initial samples generated by the frozen core model.
For its training output $I_k^{\mathrm{ret}}$, we combine image reconstruction, perceptual distance, structural similarity, and generation--prediction and generation--observation alignment under the same quotient measurement:
\begin{align}
    \mathcal L_{\mathrm{ret}}
        ={}&\lambda_{\mathrm{pix}}\mathcal L_{\mathrm{pix}}
             (I^{\mathrm{ret}},I)
          +\lambda_{\mathrm{per}}\mathcal L_{\mathrm{per}}
             (I^{\mathrm{ret}},I)
          +\lambda_{\mathrm{str}}\mathcal L_{\mathrm{str}}
             (I^{\mathrm{ret}},I) \nonumber\\
          &+\lambda_{\mathrm P}\Big\langle
             \operatorname{MSE}\!\left(
                 \mathcal M_0^b(I^{\mathrm{ret}}),\widehat C^b
             \right)\Big\rangle_b \nonumber\\
          &+\lambda_{\mathrm O}\Big\langle
             \operatorname{MSE}\!\left(
                 \mathcal M_0^b(I^{\mathrm{ret}}),\mathcal M_0^b(I)
             \right)\Big\rangle_b,
    \label{eq:qcc_return_objective}
\end{align}
where $\mathcal L_{\mathrm{pix}}$ is image MSE over the valid region, $\mathcal L_{\mathrm{per}}$ is perceptual distance, and $\mathcal L_{\mathrm{str}}$ is structural similarity loss.
The two node-level alignment terms encourage generated observations to match the observable macro-level targets and the true future observations under a shared measurement, respectively.

\section{Results}
\label{sec:results}
\subsection{Ablation Study}
\label{sec:qcc_ablation}

To examine the effect of QCC branch configurations on prediction performance, we compare four settings:
a direct action-conditioned baseline (Direct),
a single-branch QCC model with only the translation-guided branch (T-only),
a single-branch QCC model with only the rotation-guided branch (R-only),
and the complete QCC-WM with both branches.
The T-only and R-only models are trained independently, with six macro-nodes per branch.
The complete system combines dual-branch macro-state prediction, motion-guided reference transport, and spatial correction of node-level errors.
All configurations use the same evaluation split and sampling settings.
Results are reported as the mean and sample standard deviation across three training seeds.

\begin{table*}[t]
\centering
\caption{Prediction performance of different QCC configurations. Dual QCC is the complete system with macro-state prediction, motion-guided reference transport, and spatial correction of node-level errors.}
\label{tab:qcc_ablation}
\small
\setlength{\tabcolsep}{3pt}
\renewcommand{\arraystretch}{1.25}

\begin{tabular*}{\textwidth}{@{\extracolsep{\fill}}lccccc@{}}
\noalign{\hrule height 0.8pt}
\noalign{\vskip 4pt}
\textbf{Setting}
& \shortstack{ROI-MSE $\downarrow$\\$(\times 10^{-3})$}
& \shortstack{PSNR $\uparrow$\\$(\mathrm{dB})$}
& SSIM $\uparrow$
& LPIPS $\downarrow$
& \shortstack{Temporal MSE $\downarrow$\\$(\times 10^{-3})$} \\
\noalign{\vskip 4pt}
\hline
\noalign{\vskip 3pt}

Direct
& $20.46\,{\scriptstyle\pm\,0.09}$
& $24.91\,{\scriptstyle\pm\,0.01}$
& $0.7440\,{\scriptstyle\pm\,0.0003}$
& $0.0790\,{\scriptstyle\pm\,0.0006}$
& $6.17\,{\scriptstyle\pm\,0.04}$ \\

Single QCC: T-only
& $19.90\,{\scriptstyle\pm\,0.09}$
& $24.98\,{\scriptstyle\pm\,0.02}$
& $0.7450\,{\scriptstyle\pm\,0.0009}$
& $0.0782\,{\scriptstyle\pm\,0.0006}$
& $6.05\,{\scriptstyle\pm\,0.04}$ \\

Single QCC: R-only
& $19.98\,{\scriptstyle\pm\,0.10}$
& $24.98\,{\scriptstyle\pm\,0.04}$
& $0.7451\,{\scriptstyle\pm\,0.0011}$
& $0.0779\,{\scriptstyle\pm\,0.0007}$
& $6.06\,{\scriptstyle\pm\,0.05}$ \\

\noalign{\vskip 3pt}
\textbf{Dual QCC: T and R}
& $\mathbf{18.61}\,{\scriptstyle\pm\,0.08}$
& $\mathbf{25.17}\,{\scriptstyle\pm\,0.02}$
& $\mathbf{0.7492}\,{\scriptstyle\pm\,0.0001}$
& $\mathbf{0.0731}\,{\scriptstyle\pm\,0.0002}$
& $\mathbf{5.76}\,{\scriptstyle\pm\,0.02}$ \\

\noalign{\vskip 3pt}
\noalign{\hrule height 0.8pt}
\end{tabular*}

\par\vspace{4pt}
\parbox{\textwidth}{\footnotesize\raggedright
Note: Results are reported as the mean $\pm$ sample standard deviation across three training seeds.
Both MSE metrics are reported in units of $10^{-3}$.
Bold indicates the best mean in each column.}
\end{table*}

As shown in Table~\ref{tab:qcc_ablation},
both single-branch QCC variants outperform Direct in the mean values of all five metrics,
indicating that action-guided macro-level modeling improves future observation prediction.
The two single-branch configurations perform similarly overall:
T-only achieves slightly lower ROI-MSE and temporal MSE,
while R-only performs slightly better in SSIM and LPIPS.
Both achieve a PSNR of approximately $24.98\,\mathrm{dB}$.

The complete dual-branch QCC system achieves the best mean values across all five metrics.
Compared with Direct,
it reduces ROI-MSE, LPIPS, and temporal MSE by
$9.05\%$, $7.51\%$, and $6.64\%$, respectively.
It also improves all metrics over both single-branch variants,
indicating that combining dual-branch macro-state prediction with subsequent spatial correction
improves pixel reconstruction, structural similarity, perceptual quality, and prediction of inter-frame changes.

\begin{figure*}[t]
    \centering
    \includegraphics[width=0.6\textwidth]{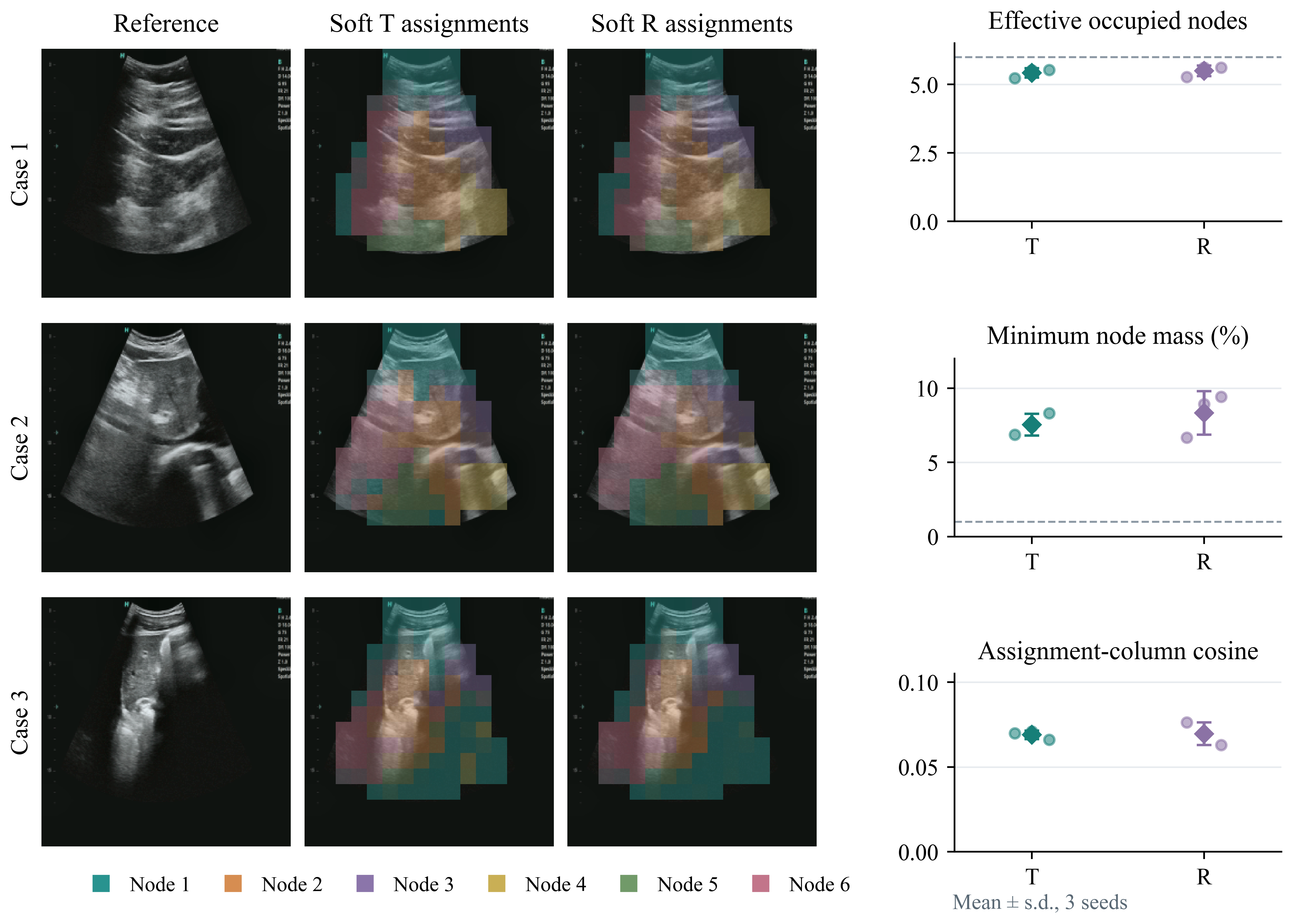}
    \caption{
    Spatial origins and effective occupancy of QCC macro-nodes.
    Left: reference observations and soft quotient assignments of the T and R branches for three samples.
    Colors indicate node indices within each branch.
    Right: effective number of occupied nodes, minimum node mass, and cosine similarity between assignment columns.
    Diamonds and error bars indicate the mean and sample standard deviation across three training seeds; circles show individual seed results.
    Dashed lines mark the six predefined nodes and the $1\%$ low-occupancy threshold, respectively.
    }
    \label{fig:qcc_node_construction}
\end{figure*}


\subsection{Mechanism Analysis of QCC-WM}
\label{sec:qcc_mechanism_analysis}

We analyze QCC from three perspectives: spatial node organization, motion prediction, and generation correction.

\subsubsection{Spatial Organization of Macro-Nodes}
\label{subsec:qcc_mechanism_construction}

QCC aggregates local visual features through action-guided soft quotient mappings while preserving the spatial support of each node in the reference observation.
The mapping remains fixed within each prediction window, ensuring that node-state comparison and subsequent spatial feedback use the same spatial correspondence.
Figure~\ref{fig:qcc_node_construction} shows that nodes in both branches cover the valid imaging region
and form distinct local aggregation regions, providing explicit spatial origins for macro-states.

We define the effective number of nodes using normalized node masses $p_j$ as
$K_{\mathrm{eff}}=\exp(-\sum_j p_j\log p_j)$.
With six nodes per branch, the effective node counts of the T and R branches are
$5.42\pm0.17$ and $5.50\pm0.19$, respectively.
Their mean minimum node masses are $(7.53\pm0.73)\%$ and $(8.34\pm1.46)\%$,
and the cosine similarities between assignment columns are $0.0689\pm0.0025$ and $0.0696\pm0.0067$.
These results indicate that node capacity is well utilized and that spatial weights have limited overlap across nodes.
This spatial organization allows macro-level motion and node-level errors to be mapped back to their corresponding local regions.

\subsubsection{Macro-Level Motion and Local Displacement}
\label{subsec:qcc_mechanism_rollout}

\begin{figure*}[t]
    \centering
    \includegraphics[width=0.6\textwidth]{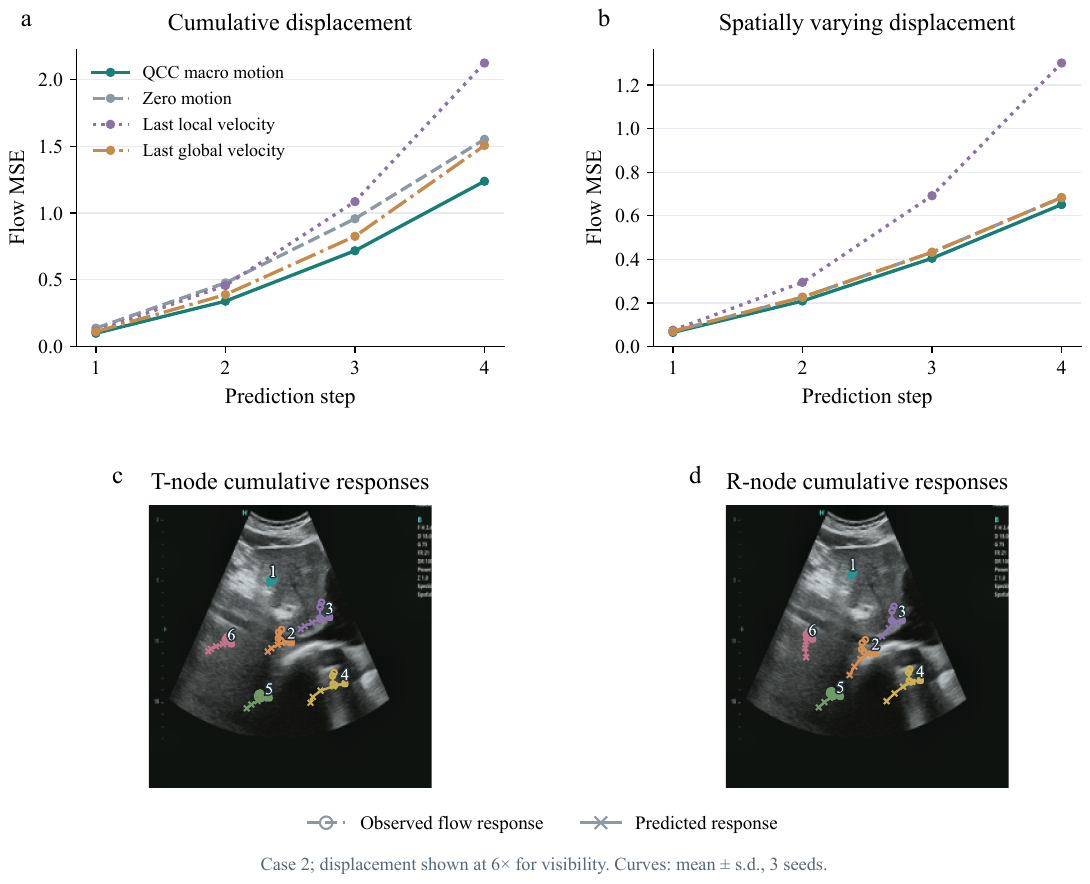}
    \caption{Macro-level motion prediction. (a) Cumulative displacement MSE. (b) MSE after removing the spatial mean of the cumulative error. Optical flow estimated from observations serves as the reference. Curves and shaded regions show the mean and sample standard deviation across three random seeds. (c,d) Cumulative responses of T and R nodes in Case 2: open circles indicate aggregated observed optical flow, and crosses indicate predictions. Successive markers correspond to consecutive prediction steps. Displacements are enlarged sixfold for visualization.}
    \label{fig:qcc_mech_rollout}
\end{figure*}

The action-conditioned dynamics predict the motion response of each node and recover local displacement fields through the reference quotient mapping.
Motion differences between macro-nodes can therefore produce different displacements across image regions, driving reference information transport.
Figure~\ref{fig:qcc_mech_rollout}(a,b) compares predicted displacements with zero-motion, local constant-velocity, and global constant-velocity extrapolation.
Using observed optical flow as the reference, we compute the MSE of two-dimensional cumulative displacement over the valid region and the corresponding error after removing its spatial mean.

QCC achieves a four-step average cumulative displacement MSE of $0.5980\pm0.0091$,
lower than the three controls at $0.7803$, $0.9451$, and $0.7071$.
After removing the spatial mean of the cumulative error, QCC achieves an MSE of $0.3321\pm0.0053$,
which remains below $0.3527$ for both zero-motion and global constant-velocity extrapolation, and $0.5906$ for local constant-velocity extrapolation.
These results indicate that macro-level dynamics reduce cumulative displacement errors and further improve the prediction of local motion differences.
Figure~\ref{fig:qcc_mech_rollout}(c,d) illustrates differences in the direction and magnitude of node responses,
showing how the fixed quotient mapping translates macro-level predictions into spatially varying motion fields.

\subsubsection{Complementary Effects of Motion Transport and Error Correction}
\label{subsec:qcc_mechanism_joint_correction}

\begin{figure*}[t]
    \centering
    \includegraphics[width=0.6\textwidth]{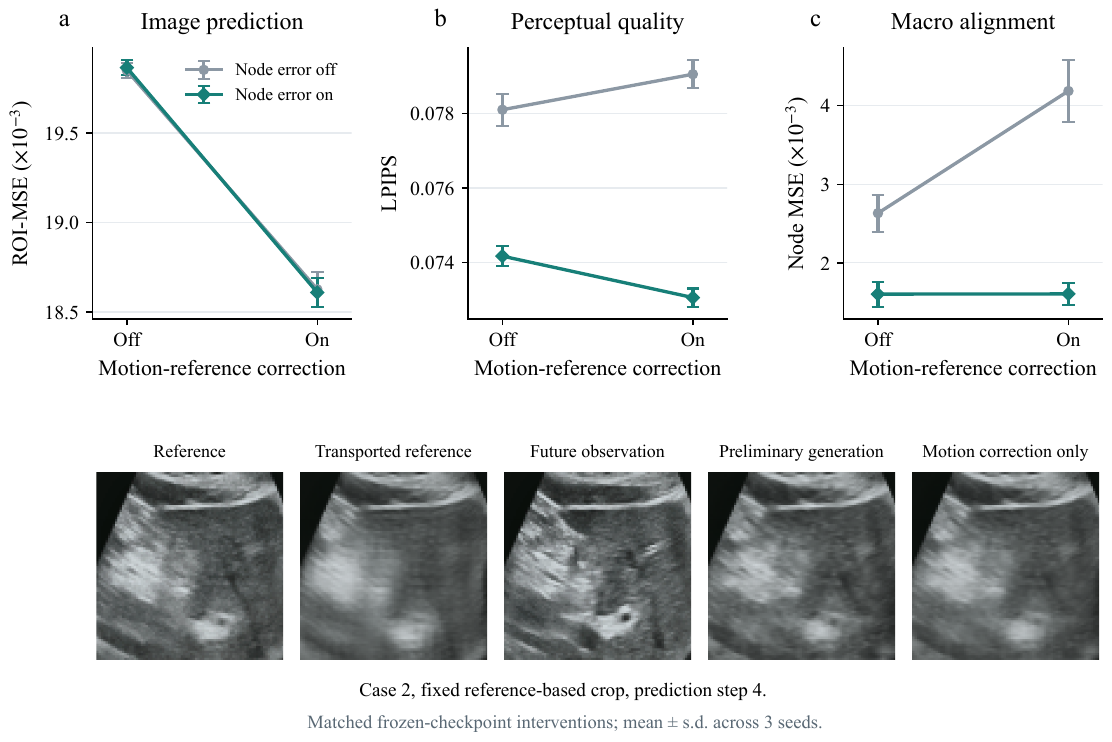}
    \caption{Complementary effects of motion transport and node-level error correction. (a--c) ROI-MSE, LPIPS, and generation--prediction node MSE under four interface settings. The horizontal axis indicates whether motion-guided reference correction is enabled. Gray and teal indicate disabled and enabled node-level error correction, respectively. Error bars show the sample standard deviation across three random seeds. Below: local crops at prediction step 4 in Case 2, showing the reference observation, transported reference, true future observation, initial generation, and generation with motion correction only.}
    \label{fig:qcc_mech_joint}
\end{figure*}

Motion-guided reference transport moves historical appearance information according to predicted displacements and corrects the initial generation through spatial gating.
Readback under the same quotient measurement then measures the remaining node-level discrepancies and maps them back to local latent space.
Figure~\ref{fig:qcc_mech_joint} examines the two paths using four settings that independently enable or disable each correction.
Node-level errors are recomputed from the generated results under each setting.

Enabling motion-guided reference correction alone reduces ROI-MSE from $0.019849\pm0.000042$
to $0.018627\pm0.000096$, indicating that transported historical appearance improves pixel reconstruction.
However, Fig.~\ref{fig:qcc_mech_joint}(b,c) shows increases in both LPIPS and generation--prediction node error.
Readback after motion transport is therefore needed to correct the remaining discrepancies from the macro-level targets.
Adding node-level error correction reduces LPIPS from $0.079051\pm0.000375$
to $0.073059\pm0.000243$, a relative reduction of $7.58\%$
(paired $95\%$ interval: $[6.39\%,8.93\%]$).
The two paths thus have complementary effects: motion transport provides spatially relevant appearance references,
while node-level error correction mainly improves perceptual quality and macro-level alignment.
Compared with disabling both corrections, the complete interface reduces ROI-MSE, LPIPS, and temporal MSE by
$6.25\%$, $6.45\%$, and $4.55\%$, respectively.

\subsubsection{Effects of Readback and Spatial Correspondence}
\label{subsec:qcc_mechanism_readback}

Readback uses the fixed reference measurement $\mathcal M_0$
to compare the observable macro-level target $\widehat C_k$ produced by motion transport with the current generation,
giving node-level errors a common measurement definition and spatial origin.
To evaluate correction, we compute node-level errors between generated observations and both macro-level targets and true future observations:
\begin{align}
    E_{\mathrm{G\text{-}P}}(k)
    &=\operatorname{MSE}\!\left(
        \mathcal M_0(\widehat U_k),\widehat C_k
    \right), \label{eq:qcc_mech_generation_prediction}\\
    E_{\mathrm{G\text{-}O}}(k)
    &=\operatorname{MSE}\!\left(
        \mathcal M_0(\widehat U_k),\mathcal M_0(U_k)
    \right), \label{eq:qcc_mech_generation_observation}
\end{align}
where $\widehat U_k$ and $U_k$ denote generated and true future observations, respectively, and the metrics are averaged over the T and R branches.

\begin{figure*}[t]
    \centering
    \includegraphics[width=0.6\textwidth]{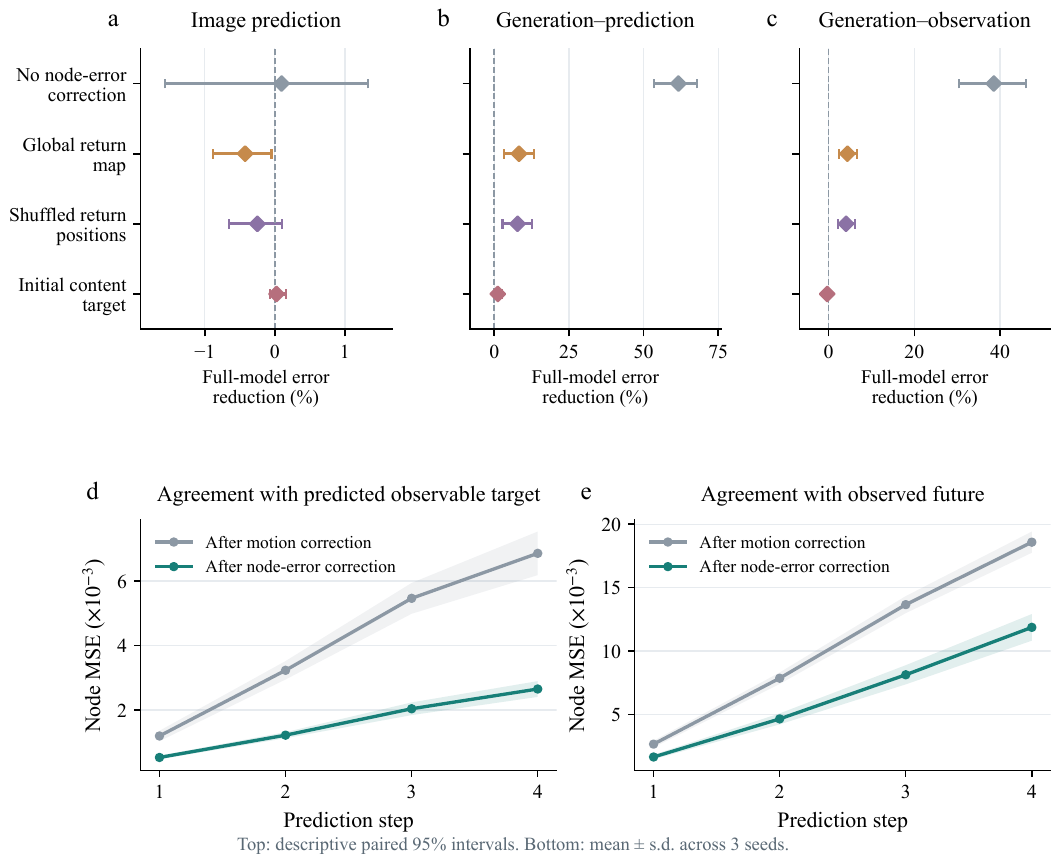}
    \caption{Analysis of readback under the same quotient measurement and spatial feedback. (a--c) Reductions in image error, generation--prediction node error, and generation--observation node error for the complete model relative to four controls: no node-level error correction, global feedback, spatially shuffled feedback, and initial content targets. Positive values favor the complete model; horizontal intervals indicate paired $95\%$ intervals. (d,e) Node MSE at each prediction step after motion correction and after adding node-level error correction. Curves and shaded regions show the mean and sample standard deviation across three random seeds. All settings use the same reference quotient measurement.}
    \label{fig:qcc_mech_readback}
\end{figure*}

Adding node-level error feedback reduces generation--prediction node MSE from $0.004187$ to $0.001604$
and generation--observation node MSE from $0.010688$ to $0.006570$,
corresponding to reductions of $61.69\%$ and $38.53\%$
(paired $95\%$ intervals: $[53.36\%,67.91\%]$ and $[30.34\%,46.14\%]$, respectively).
Figure~\ref{fig:qcc_mech_readback}(d,e) shows that both errors decrease at all four prediction steps.
Readback correction therefore improves both agreement with macro-level predictions and macro-level correspondence with true future observations.

To examine the effect of error feedback locations, we replace the final feedback mapping with a spatially uniform assignment or shuffle its spatial positions,
while keeping all other predictions, readback measurements, and correction conditions unchanged.
Global feedback preserves the valid support and total node masses.
Results for spatial shuffling are averaged over five predefined shuffles.
Compared with these two controls, complete spatial feedback reduces generation--prediction node MSE by $8.28\%$ and $7.79\%$,
and generation--observation node MSE by $4.39\%$ and $4.07\%$, respectively.
ROI-MSE values for global, shuffled, and complete feedback are $0.018530$, $0.018563$, and $0.018609$, respectively.
Complete feedback has slightly higher pixel error, with its main advantage lying in macro-level alignment.
These results indicate that preserving node--location correspondence helps translate macro-level errors into generation adjustments at the appropriate regions.

Compared with using the initial content measurement as the error decoder target, using the future observable target reduces generation--prediction node error by $1.20\%$.
ROI-MSE changes by only $0.02\%$, with its paired interval including zero.
Together, these results suggest that the current correction gains mainly arise from combining motion-guided reference transport with node-level error feedback,
while correct spatial correspondence further improves consistency between macro-level predictions and generated observations.

\subsection{Comparative Experiments}
\label{sec:horizon_protocol}

We evaluate action-conditioned ultrasound video prediction through short-window prediction and multi-step extrapolation,
with prediction horizons $H\in\{4,8,16\}$.
All methods use model weights trained at $H=4$ and recursively generate future observations in blocks of four frames.
The $H=8$ and $H=16$ settings correspond to extrapolation over two and four prediction blocks, respectively.
Each model starts from four real historical frames.
Each prediction block receives the corresponding four-step sequence of six-dimensional actions,
and generated observations are used as history for the next block.
The $H=8$ results are taken from the first eight steps of the same sixteen-step trajectory to preserve the same noise and prediction prefix.

The compared methods include PredRNN-Act~\cite{wang2017predrnn,wang2023predrnnv2},
PredRNN-V2-Act~\cite{wang2023predrnnv2}, SimVP~\cite{gao2022simvp},
IRASim-S/4~\cite{zhu2025irasim}, DINO-WM~\cite{zhou2025dinowm},
MAU~\cite{chang2021mau}, and MMVP~\cite{zhong2023mmvp}.
The PredRNN models use the action-conditioned extensions released by their authors.
SimVP, MAU, and MMVP use implementations from OpenSTL~\cite{tan2023openstl}.
Each baseline retains its original backbone, with separate interfaces adapting action inputs and observation representations.
The PredRNN models, SimVP, IRASim, MAU, and MMVP share the same frozen domain-specific VAE.
DINO-WM uses its native DINOv2 feature dynamics and official RGB decoder.
During recursive prediction, QCC feeds back the corrected latent variables and recomputes USFM features, JEPA features, and historical optical flow from generated frames to construct quotient states for the next prediction block.
DINO-WM feeds back predicted patch features, while the remaining methods feed back predicted latent variables.
All baselines are trained for 30 epochs and evaluated using the raw final-epoch weights.

Based on recorded timestamps, the median temporal spans for $H=4,8,16$ are approximately
$0.133\,\mathrm{s}$, $0.267\,\mathrm{s}$, and $0.533\,\mathrm{s}$, respectively.

\subsubsection{Four-Step Prediction}
\label{subsec:horizon_four}

Table~\ref{tab:horizon_4} reports four-step prediction results.
QCC achieves the best mean values across all five metrics.
Compared with SimVP, the baseline with the lowest ROI-MSE, QCC reduces ROI-MSE by 2.70\% and increases PSNR by $0.862\,\mathrm{dB}$.
Compared with MMVP, the strongest baseline in perceptual quality and temporal prediction,
QCC reduces LPIPS and temporal MSE by 47.84\% and 13.06\%, respectively.
These results indicate improvements in pixel reconstruction, perceptual quality, and prediction of inter-frame changes within the short prediction window,
with the largest relative reduction observed in perceptual distance.

\begin{table*}[!t]
\centering
\caption{Cumulative prediction results at $H=4$.}
\label{tab:horizon_4}
\small
\setlength{\tabcolsep}{3pt}
\renewcommand{\arraystretch}{1.25}
\begin{tabular*}{\textwidth}{@{\extracolsep{\fill}}lccccc@{}}
\noalign{\hrule height 0.8pt}
\noalign{\vskip 4pt}
\textbf{Method} & \shortstack{ROI-MSE $\downarrow$\\$(\times 10^{-3})$} & \shortstack{PSNR $\uparrow$\\$(\mathrm{dB})$} & SSIM $\uparrow$ & LPIPS $\downarrow$ & \shortstack{Temporal MSE $\downarrow$\\$(\times 10^{-3})$} \\
\noalign{\vskip 4pt}
\hline
\noalign{\vskip 3pt}
PredRNN-Act~\cite{wang2017predrnn} & $24.51\,{\scriptstyle\pm\,0.76}$ & $23.22\,{\scriptstyle\pm\,0.11}$ & $0.7053\,{\scriptstyle\pm\,0.0009}$ & $0.1530\,{\scriptstyle\pm\,0.0013}$ & $6.92\,{\scriptstyle\pm\,0.09}$ \\
PredRNN-V2-Act~\cite{wang2023predrnnv2} & $23.99\,{\scriptstyle\pm\,0.81}$ & $23.30\,{\scriptstyle\pm\,0.12}$ & $0.7025\,{\scriptstyle\pm\,0.0017}$ & $0.1515\,{\scriptstyle\pm\,0.0008}$ & $6.85\,{\scriptstyle\pm\,0.09}$ \\
SimVP+Act~\cite{gao2022simvp} & $19.70\,{\scriptstyle\pm\,0.24}$ & $24.14\,{\scriptstyle\pm\,0.07}$ & $0.7251\,{\scriptstyle\pm\,0.0022}$ & $0.2110\,{\scriptstyle\pm\,0.0020}$ & $7.35\,{\scriptstyle\pm\,0.08}$ \\
IRASim-S/4~\cite{zhu2025irasim} & $48.67\,{\scriptstyle\pm\,5.19}$ & $20.28\,{\scriptstyle\pm\,0.38}$ & $0.6152\,{\scriptstyle\pm\,0.0429}$ & $0.1677\,{\scriptstyle\pm\,0.0338}$ & $28.43\,{\scriptstyle\pm\,8.21}$ \\
DINO-WM+Dec.~\cite{zhou2025dinowm} & $37.84\,{\scriptstyle\pm\,2.37}$ & $20.53\,{\scriptstyle\pm\,0.28}$ & $0.5977\,{\scriptstyle\pm\,0.0025}$ & $0.2879\,{\scriptstyle\pm\,0.0417}$ & $12.08\,{\scriptstyle\pm\,0.61}$ \\
MAU+Act~\cite{chang2021mau} & $23.54\,{\scriptstyle\pm\,0.20}$ & $23.11\,{\scriptstyle\pm\,0.09}$ & $0.6964\,{\scriptstyle\pm\,0.0061}$ & $0.2095\,{\scriptstyle\pm\,0.0037}$ & $7.29\,{\scriptstyle\pm\,0.16}$ \\
MMVP+Act~\cite{zhong2023mmvp} & $22.98\,{\scriptstyle\pm\,0.25}$ & $23.93\,{\scriptstyle\pm\,0.04}$ & $0.7294\,{\scriptstyle\pm\,0.0024}$ & $0.1421\,{\scriptstyle\pm\,0.0070}$ & $6.73\,{\scriptstyle\pm\,0.04}$ \\
\noalign{\vskip 3pt}
\textbf{QCC-WM} & $\mathbf{18.61}\,{\scriptstyle\pm\,0.08}$
& $\mathbf{25.17}\,{\scriptstyle\pm\,0.02}$
& $\mathbf{0.7492}\,{\scriptstyle\pm\,0.0001}$
& $\mathbf{0.0731}\,{\scriptstyle\pm\,0.0002}$
& $\mathbf{5.76}\,{\scriptstyle\pm\,0.02}$ \\
\noalign{\vskip 3pt}
\noalign{\hrule height 0.8pt}
\end{tabular*}
\par\vspace{4pt}
\parbox{\textwidth}{\footnotesize\raggedright
Note: All methods use the same 226 test starting points. Results are the mean $\pm$ sample standard deviation across three training seeds.
Both MSE metrics are reported in units of $10^{-3}$; bold indicates the best mean.
Act denotes an action-conditioning interface, and Dec. denotes the official RGB decoder.}
\end{table*}

\subsubsection{Eight-Step Extrapolation}
\label{subsec:horizon_eight}

Table~\ref{tab:horizon_8} reports results after recursively generating two prediction blocks.
QCC maintains the best mean values in ROI-MSE, PSNR, LPIPS, and temporal MSE.
Compared with the best baseline for each metric, QCC reduces ROI-MSE by 2.05\% relative to SimVP,
and reduces LPIPS and temporal MSE by 41.68\% and 5.65\%, respectively, relative to PredRNN-V2.
QCC achieves an SSIM of 0.64475, close to MMVP at 0.64526.
Reconstruction errors and perceptual distances increase for all methods compared with four-step prediction.
QCC nevertheless retains a substantial LPIPS reduction beyond the training window, indicating a perceptual quality advantage in multi-step extrapolation.

\begin{table*}[!t]
\centering
\caption{Cumulative prediction results at $H=8$.}
\label{tab:horizon_8}
\small
\setlength{\tabcolsep}{3pt}
\renewcommand{\arraystretch}{1.25}
\begin{tabular*}{\textwidth}{@{\extracolsep{\fill}}lccccc@{}}
\noalign{\hrule height 0.8pt}
\noalign{\vskip 4pt}
\textbf{Method} & \shortstack{ROI-MSE $\downarrow$\\$(\times 10^{-3})$} & \shortstack{PSNR $\uparrow$\\$(\mathrm{dB})$} & SSIM $\uparrow$ & LPIPS $\downarrow$ & \shortstack{Temporal MSE $\downarrow$\\$(\times 10^{-3})$} \\
\noalign{\vskip 4pt}
\hline
\noalign{\vskip 3pt}
PredRNN-Act~\cite{wang2017predrnn} & $40.33\,{\scriptstyle\pm\,1.44}$ & $21.04\,{\scriptstyle\pm\,0.15}$ & $0.6264\,{\scriptstyle\pm\,0.0018}$ & $0.1958\,{\scriptstyle\pm\,0.0022}$ & $7.51\,{\scriptstyle\pm\,0.06}$ \\
PredRNN-V2-Act~\cite{wang2023predrnnv2} & $39.96\,{\scriptstyle\pm\,1.73}$ & $21.09\,{\scriptstyle\pm\,0.16}$ & $0.6226\,{\scriptstyle\pm\,0.0033}$ & $0.1913\,{\scriptstyle\pm\,0.0009}$ & $7.43\,{\scriptstyle\pm\,0.06}$ \\
SimVP+Act~\cite{gao2022simvp} & $35.86\,{\scriptstyle\pm\,0.58}$ & $21.68\,{\scriptstyle\pm\,0.08}$ & $0.6414\,{\scriptstyle\pm\,0.0018}$ & $0.2542\,{\scriptstyle\pm\,0.0019}$ & $8.27\,{\scriptstyle\pm\,0.09}$ \\
IRASim-S/4~\cite{zhu2025irasim} & $85.89\,{\scriptstyle\pm\,17.20}$ & $17.97\,{\scriptstyle\pm\,0.63}$ & $0.5151\,{\scriptstyle\pm\,0.0354}$ & $0.2147\,{\scriptstyle\pm\,0.0288}$ & $28.45\,{\scriptstyle\pm\,7.41}$ \\
DINO-WM+Dec.~\cite{zhou2025dinowm} & $50.65\,{\scriptstyle\pm\,3.57}$ & $19.40\,{\scriptstyle\pm\,0.30}$ & $0.5601\,{\scriptstyle\pm\,0.0041}$ & $0.2978\,{\scriptstyle\pm\,0.0403}$ & $10.17\,{\scriptstyle\pm\,0.36}$ \\
MAU+Act~\cite{chang2021mau} & $40.11\,{\scriptstyle\pm\,0.81}$ & $20.94\,{\scriptstyle\pm\,0.07}$ & $0.6199\,{\scriptstyle\pm\,0.0060}$ & $0.2363\,{\scriptstyle\pm\,0.0048}$ & $7.85\,{\scriptstyle\pm\,0.17}$ \\
MMVP+Act~\cite{zhong2023mmvp} & $38.10\,{\scriptstyle\pm\,0.32}$ & $21.57\,{\scriptstyle\pm\,0.08}$ & $\mathbf{0.6453}\,{\scriptstyle\pm\,0.0032}$ & $0.2058\,{\scriptstyle\pm\,0.0087}$ & $7.54\,{\scriptstyle\pm\,0.06}$ \\
\noalign{\vskip 3pt}
\textbf{QCC-WM} & $\mathbf{35.12}\,{\scriptstyle\pm\,0.30}$ & $\mathbf{22.26}\,{\scriptstyle\pm\,0.05}$ & $0.6448\,{\scriptstyle\pm\,0.0007}$ & $\mathbf{0.1116}\,{\scriptstyle\pm\,0.0003}$ & $\mathbf{7.02}\,{\scriptstyle\pm\,0.02}$ \\
\noalign{\vskip 3pt}
\noalign{\hrule height 0.8pt}
\end{tabular*}
\par\vspace{4pt}
\parbox{\textwidth}{\footnotesize\raggedright
Note: All methods use the same 226 test starting points. Results are the mean $\pm$ sample standard deviation across three training seeds.
Both MSE metrics are reported in units of $10^{-3}$; bold indicates the best mean.
Act denotes an action-conditioning interface, and Dec. denotes the official RGB decoder.}
\end{table*}

\subsubsection{Sixteen-Step Extrapolation}
\label{subsec:horizon_sixteen}

Table~\ref{tab:horizon_16} reports results after recursively generating four prediction blocks.
QCC retains the best mean values in ROI-MSE, PSNR, LPIPS, and temporal MSE.
Its LPIPS is 0.16209, a 30.23\% reduction relative to the best baseline, PredRNN-V2.
QCC achieves an ROI-MSE of 0.056069, which is 0.77\% lower than MMVP at 0.056504,
and a temporal MSE of 0.007315, which is 1.55\% lower than PredRNN-V2.
As the extrapolation horizon increases, QCC's relative advantages in pixel and temporal errors decrease, while its advantage in perceptual distance remains substantial.
For SSIM, MMVP achieves the best result at 0.56690.
QCC achieves 0.55178, below MMVP, SimVP, and PredRNN-Act, indicating that these methods provide higher structural similarity at this horizon.

\begin{table*}[!t]
\centering
\caption{Cumulative prediction results at $H=16$.}
\label{tab:horizon_16}
\small
\setlength{\tabcolsep}{3pt}
\renewcommand{\arraystretch}{1.25}
\begin{tabular*}{\textwidth}{@{\extracolsep{\fill}}lccccc@{}}
\noalign{\hrule height 0.8pt}
\noalign{\vskip 4pt}
\textbf{Method} & \shortstack{ROI-MSE $\downarrow$\\$(\times 10^{-3})$} & \shortstack{PSNR $\uparrow$\\$(\mathrm{dB})$} & SSIM $\uparrow$ & LPIPS $\downarrow$ & \shortstack{Temporal MSE $\downarrow$\\$(\times 10^{-3})$} \\
\noalign{\vskip 4pt}
\hline
\noalign{\vskip 3pt}
PredRNN-Act~\cite{wang2017predrnn} & $58.05\,{\scriptstyle\pm\,2.28}$ & $19.23\,{\scriptstyle\pm\,0.18}$ & $0.5553\,{\scriptstyle\pm\,0.0020}$ & $0.2407\,{\scriptstyle\pm\,0.0037}$ & $7.51\,{\scriptstyle\pm\,0.04}$ \\
PredRNN-V2-Act~\cite{wang2023predrnnv2} & $58.17\,{\scriptstyle\pm\,2.16}$ & $19.25\,{\scriptstyle\pm\,0.16}$ & $0.5506\,{\scriptstyle\pm\,0.0029}$ & $0.2323\,{\scriptstyle\pm\,0.0015}$ & $7.43\,{\scriptstyle\pm\,0.04}$ \\
SimVP+Act~\cite{gao2022simvp} & $56.79\,{\scriptstyle\pm\,1.31}$ & $19.50\,{\scriptstyle\pm\,0.10}$ & $0.5633\,{\scriptstyle\pm\,0.0024}$ & $0.2930\,{\scriptstyle\pm\,0.0025}$ & $8.34\,{\scriptstyle\pm\,0.04}$ \\
IRASim-S/4~\cite{zhu2025irasim} & $147.50\,{\scriptstyle\pm\,50.71}$ & $15.74\,{\scriptstyle\pm\,1.16}$ & $0.4200\,{\scriptstyle\pm\,0.0285}$ & $0.2678\,{\scriptstyle\pm\,0.0233}$ & $28.87\,{\scriptstyle\pm\,7.06}$ \\
DINO-WM+Dec.~\cite{zhou2025dinowm} & $69.31\,{\scriptstyle\pm\,5.44}$ & $18.10\,{\scriptstyle\pm\,0.33}$ & $0.5121\,{\scriptstyle\pm\,0.0101}$ & $0.3134\,{\scriptstyle\pm\,0.0390}$ & $8.82\,{\scriptstyle\pm\,0.21}$ \\
MAU+Act~\cite{chang2021mau} & $64.12\,{\scriptstyle\pm\,2.86}$ & $18.87\,{\scriptstyle\pm\,0.15}$ & $0.5423\,{\scriptstyle\pm\,0.0082}$ & $0.2668\,{\scriptstyle\pm\,0.0060}$ & $7.81\,{\scriptstyle\pm\,0.19}$ \\
MMVP+Act~\cite{zhong2023mmvp} & $56.50\,{\scriptstyle\pm\,1.53}$ & $19.53\,{\scriptstyle\pm\,0.13}$ & $\mathbf{0.5669}\,{\scriptstyle\pm\,0.0051}$ & $0.2715\,{\scriptstyle\pm\,0.0059}$ & $7.61\,{\scriptstyle\pm\,0.08}$ \\
\noalign{\vskip 3pt}
\textbf{QCC-WM} & $\mathbf{56.07}\,{\scriptstyle\pm\,0.07}$ & $\mathbf{19.86}\,{\scriptstyle\pm\,0.02}$ & $0.5518\,{\scriptstyle\pm\,0.0010}$ & $\mathbf{0.1621}\,{\scriptstyle\pm\,0.0007}$ & $\mathbf{7.32}\,{\scriptstyle\pm\,0.02}$ \\
\noalign{\vskip 3pt}
\noalign{\hrule height 0.8pt}
\end{tabular*}
\par\vspace{4pt}
\parbox{\textwidth}{\footnotesize\raggedright
Note: All methods use the same 226 test starting points. Results are the mean $\pm$ sample standard deviation across three training seeds.
Both MSE metrics are reported in units of $10^{-3}$; bold indicates the best mean.
Act denotes an action-conditioning interface, and Dec. denotes the official RGB decoder.}
\end{table*}

Across the three prediction horizons, QCC achieves the best mean values for all five metrics at four steps
and maintains its lead in ROI-MSE, PSNR, LPIPS, and temporal MSE at eight and sixteen steps.
Its LPIPS reductions relative to the best baseline at each horizon are 47.84\%, 41.68\%, and 30.23\%, respectively,
representing its most prominent performance advantage across prediction horizons.

As the prediction horizon increases from 4 to 16 steps, QCC's cumulative ROI-MSE rises from 0.01861 to 0.05606,
and LPIPS increases from 0.0731 to 0.1620, approximately 2.92 and 2.19 times their four-step values, respectively.
PSNR decreases by $5.145\,\mathrm{dB}$, and SSIM decreases by 0.19010.
This trend reflects error accumulation during recursive prediction.
Overall, QCC maintains favorable perceptual quality during multi-step extrapolation,
while structural similarity at longer horizons remains an area for improvement.

\subsection{Experiments on an Additional Dataset}

To evaluate the transfer performance of QCC-WM on additional ultrasound data, we conduct supplementary experiments on the public
Freehand US dataset~\cite{li2022freehand,li2023trackerless}.
The dataset contains ultrasound scans of both forearms from 19 volunteers, with 12 scanning sequences per volunteer and 228 sequences in total.
Acquisition covers straight, C-shaped, and S-shaped trajectories, with the probe oriented parallel or perpendicular to the forearm.
The B-mode images have an original resolution of $480\times640$ and a sampling rate of approximately $20\,\mathrm{Hz}$.
The released data also include optically tracked poses and calibration information.
We represent calibrated relative $\mathrm{SE}(3)$ transformations between adjacent frames using three-dimensional translation and three-dimensional axis-angle rotation,
forming six-dimensional action conditions for predicting subsequent ultrasound images.

The training, validation, and test sets are split by subject and contain 13, 3, and 3 subjects, respectively.
The training set provides 1437 four-step prediction windows.
The validation set contains 207 windows with at least 16 consecutive future steps, and the test set contains 278 common prediction starting points.
Images are resized to $224\times224$ and normalized to $[-1,1]$.
We retain the original temporal sampling and align images with poses using the official calibration information.
All methods receive four historical frames and the corresponding action sequences and are trained with a supervision horizon of $H=4$.
During evaluation, the same $H=4$ model weights are reused to recursively generate predictions at $H=8$ and $H=16$ in four-frame blocks,
with generated observations fed back between blocks.
All three horizons are evaluated at the same 278 test starting points.
Their median temporal spans are approximately $0.196$, $0.391$, and $0.798\,\mathrm{s}$, respectively.

Table~\ref{tab:freehand_results} reports the comparative results on Freehand US.
QCC-WM achieves the best mean values across all five metrics at all three horizons.
At $H=4$, QCC-WM reduces ROI-MSE by 33.09\% and increases PSNR by $2.101\,\mathrm{dB}$ relative to SimVP-Act, the baseline with the lowest pixel error.
It reduces LPIPS by 23.57\% relative to PredRNN-Act, the baseline with the lowest perceptual distance.
These results indicate that QCC-WM improves pixel reconstruction, structural similarity, and perceptual prediction quality in this transfer setting.

As the prediction horizon increases, ROI-MSE generally rises and SSIM decreases across methods,
reflecting error accumulation during recursive prediction.
At $H=8$ and $H=16$, QCC-WM still reduces ROI-MSE by 18.24\% and 5.88\%, respectively, relative to the best baseline at each horizon.
At $H=16$, QCC-WM achieves an SSIM of $0.7788$, exceeding SimVP-Act at $0.7698$,
and reduces temporal MSE by 8.17\% relative to PredRNN-V2-Act, the best baseline for this metric.
Although the relative pixel-error advantage narrows as the extrapolation horizon increases,
QCC-WM maintains favorable structural and temporal prediction performance on this additional dataset.

\begin{table*}[htbp]
\centering
\caption{Action-conditioned ultrasound video prediction on Freehand US.
All methods use the same 278 test starting points. Results are the mean $\pm$ sample standard deviation across three random seeds.
At $H=8,16$, models trained for four-step prediction are reused for longer-horizon extrapolation.
ROI-MSE and temporal MSE are reported in units of $10^{-3}$. Bold indicates the best mean at each prediction horizon.}
\label{tab:freehand_results}
\small
\setlength{\tabcolsep}{3.5pt}
\renewcommand{\arraystretch}{1.13}
\resizebox{\linewidth}{!}{%
\begin{tabular}{clccccc}
\toprule
Horizon & Method & ROI-MSE $\downarrow$ & PSNR/dB $\uparrow$ & SSIM $\uparrow$ & LPIPS $\downarrow$ & Temporal MSE $\downarrow$ \\
\midrule

\multirow{8}{*}{$H=4$}
 & PredRNN-Act~\cite{wang2017predrnn}
 & $4.382\!\pm\!0.150$
 & $29.775\!\pm\!0.159$
 & $0.8278\!\pm\!0.0007$
 & $0.1448\!\pm\!0.0014$
 & $1.553\!\pm\!0.014$ \\

 & PredRNN-V2-Act~\cite{wang2023predrnnv2}
 & $4.524\!\pm\!0.118$
 & $29.642\!\pm\!0.107$
 & $0.8283\!\pm\!0.0007$
 & $0.1475\!\pm\!0.0008$
 & $1.561\!\pm\!0.009$ \\

 & SimVP-Act~\cite{gao2022simvp}
 & $4.056\!\pm\!0.015$
 & $30.048\!\pm\!0.015$
 & $0.8408\!\pm\!0.0004$
 & $0.1933\!\pm\!0.0008$
 & $1.663\!\pm\!0.007$ \\

 & IRASim-S/4~\cite{zhu2025irasim}
 & $20.237\!\pm\!9.006$
 & $24.285\!\pm\!2.010$
 & $0.6325\!\pm\!0.0457$
 & $0.2130\!\pm\!0.0292$
 & $14.124\!\pm\!4.614$ \\

 & DINO-WM~\cite{zhou2025dinowm}
 & $8.148\!\pm\!0.509$
 & $27.220\!\pm\!0.208$
 & $0.7527\!\pm\!0.0010$
 & $0.2500\!\pm\!0.0061$
 & $2.631\!\pm\!0.069$ \\

 & MAU-Act~\cite{chang2021mau}
 & $4.203\!\pm\!0.182$
 & $29.922\!\pm\!0.187$
 & $0.8307\!\pm\!0.0013$
 & $0.2048\!\pm\!0.0010$
 & $1.581\!\pm\!0.016$ \\

 & MMVP-Act~\cite{zhong2023mmvp}
 & $4.139\!\pm\!0.093$
 & $30.039\!\pm\!0.099$
 & $0.8450\!\pm\!0.0035$
 & $0.1469\!\pm\!0.0113$
 & $1.522\!\pm\!0.050$ \\

 & \textbf{QCC-WM}
 & $\mathbf{2.714}\!\pm\!\mathbf{0.014}$
 & $\mathbf{32.149}\!\pm\!\mathbf{0.021}$
 & $\mathbf{0.8626}\!\pm\!\mathbf{0.0002}$
 & $\mathbf{0.1107}\!\pm\!\mathbf{0.0009}$
 & $\mathbf{1.132}\!\pm\!\mathbf{0.003}$ \\

\midrule

\multirow{8}{*}{$H=8$}
 & PredRNN-Act~\cite{wang2017predrnn}
 & $6.080\!\pm\!0.099$
 & $28.550\!\pm\!0.094$
 & $0.7896\!\pm\!0.0004$
 & $0.1777\!\pm\!0.0013$
 & $1.315\!\pm\!0.005$ \\

 & PredRNN-V2-Act~\cite{wang2023predrnnv2}
 & $6.263\!\pm\!0.241$
 & $28.416\!\pm\!0.154$
 & $0.7900\!\pm\!0.0003$
 & $0.1810\!\pm\!0.0010$
 & $1.315\!\pm\!0.004$ \\

 & SimVP-Act~\cite{gao2022simvp}
 & $5.385\!\pm\!0.039$
 & $28.999\!\pm\!0.022$
 & $0.8106\!\pm\!0.0003$
 & $0.2205\!\pm\!0.0008$
 & $1.402\!\pm\!0.004$ \\

 & IRASim-S/4~\cite{zhu2025irasim}
 & $35.127\!\pm\!19.700$
 & $22.339\!\pm\!2.483$
 & $0.5707\!\pm\!0.0577$
 & $0.2482\!\pm\!0.0329$
 & $13.297\!\pm\!4.581$ \\

 & DINO-WM~\cite{zhou2025dinowm}
 & $10.352\!\pm\!0.792$
 & $26.312\!\pm\!0.252$
 & $0.7360\!\pm\!0.0006$
 & $0.2571\!\pm\!0.0063$
 & $1.894\!\pm\!0.035$ \\

 & MAU-Act~\cite{chang2021mau}
 & $5.797\!\pm\!0.309$
 & $28.728\!\pm\!0.230$
 & $0.7979\!\pm\!0.0022$
 & $0.2238\!\pm\!0.0018$
 & $1.339\!\pm\!0.007$ \\

 & MMVP-Act~\cite{zhong2023mmvp}
 & $5.702\!\pm\!0.143$
 & $28.851\!\pm\!0.108$
 & $0.8099\!\pm\!0.0036$
 & $0.1883\!\pm\!0.0121$
 & $1.303\!\pm\!0.037$ \\

 & \textbf{QCC-WM}
 & $\mathbf{4.403}\!\pm\!\mathbf{0.005}$
 & $\mathbf{30.413}\!\pm\!\mathbf{0.011}$
 & $\mathbf{0.8251}\!\pm\!\mathbf{0.0004}$
 & $\mathbf{0.1391}\!\pm\!\mathbf{0.0015}$
 & $\mathbf{1.106}\!\pm\!\mathbf{0.001}$ \\

\midrule

\multirow{8}{*}{$H=16$}
 & PredRNN-Act~\cite{wang2017predrnn}
 & $9.424\!\pm\!0.632$
 & $26.934\!\pm\!0.168$
 & $0.7447\!\pm\!0.0010$
 & $0.2182\!\pm\!0.0009$
 & $1.196\!\pm\!0.008$ \\

 & PredRNN-V2-Act~\cite{wang2023predrnnv2}
 & $9.533\!\pm\!0.569$
 & $26.841\!\pm\!0.229$
 & $0.7442\!\pm\!0.0014$
 & $0.2206\!\pm\!0.0013$
 & $1.193\!\pm\!0.003$ \\

 & SimVP-Act~\cite{gao2022simvp}
 & $8.260\!\pm\!0.102$
 & $27.458\!\pm\!0.039$
 & $0.7698\!\pm\!0.0007$
 & $0.2500\!\pm\!0.0012$
 & $1.276\!\pm\!0.003$ \\

 & IRASim-S/4~\cite{zhu2025irasim}
 & $69.113\!\pm\!51.126$
 & $20.035\!\pm\!3.209$
 & $0.4915\!\pm\!0.0785$
 & $0.2999\!\pm\!0.0387$
 & $13.449\!\pm\!5.119$ \\

 & DINO-WM~\cite{zhou2025dinowm}
 & $15.334\!\pm\!1.459$
 & $24.849\!\pm\!0.319$
 & $0.7081\!\pm\!0.0021$
 & $0.2709\!\pm\!0.0070$
 & $1.508\!\pm\!0.021$ \\

 & MAU-Act~\cite{chang2021mau}
 & $9.480\!\pm\!0.691$
 & $26.950\!\pm\!0.291$
 & $0.7532\!\pm\!0.0054$
 & $0.2483\!\pm\!0.0045$
 & $1.223\!\pm\!0.008$ \\

 & MMVP-Act~\cite{zhong2023mmvp}
 & $8.550\!\pm\!0.249$
 & $27.348\!\pm\!0.129$
 & $0.7677\!\pm\!0.0040$
 & $0.2344\!\pm\!0.0096$
 & $1.194\!\pm\!0.025$ \\

 & \textbf{QCC-WM}
 & $\mathbf{7.775}\!\pm\!\mathbf{0.102}$
 & $\mathbf{28.349}\!\pm\!\mathbf{0.024}$
 & $\mathbf{0.7788}\!\pm\!\mathbf{0.0003}$
 & $\mathbf{0.1852}\!\pm\!\mathbf{0.0031}$
 & $\mathbf{1.095}\!\pm\!\mathbf{0.001}$ \\

\bottomrule
\end{tabular}%
}
\end{table*}

\section{Conclusion}
\label{sec:conclusion}

We presented QCC-WM, an action-conditioned ultrasound video prediction model that constructs macro-states with spatial correspondence through action-guided quotient mappings. By combining recursive state prediction, motion-guided reference transport, and readback correction under the same quotient measurement, QCC-WM connects macro-level dynamics with future observation generation. Comparative experiments on US-VLA and transfer experiments on Freehand US show improved ultrasound video prediction quality and a sustained perceptual advantage during recursive extrapolation beyond the training window. Ablation studies and mechanism analyses further show that motion-guided reference transport mainly improves pixel reconstruction, while node-level error correction further improves perceptual quality and consistency between generated observations and macro-level predictions. Future work will focus on improving structural preservation in long-horizon prediction and evaluating the framework for closed-loop ultrasound navigation.

\bibliographystyle{elsarticle-num}
\bibliography{references}
\end{document}